# Modified Gaussian Process Regression Models for Cyclic Capacity Prediction of Lithium-ion Batteries

Kailong Liu, *Member, IEEE*, Xiaosong Hu, *Senior Member, IEEE*, Zhongbao Wei, *Member, IEEE,* Yi Li, and Yan Jiang

*Abstract*—This paper presents the development of machine learning-enabled data-driven models for effective capacity predictions for lithium-ion batteries under different cyclic conditions. To achieve this, a model structure is first proposed with the considerations of battery ageing tendency and the corresponding operational temperature and depth-of-discharge. Then based on a systematic understanding of covariance functions within the Gaussian process regression, two related data-driven models are developed. Specifically, by modifying the isotropic squared exponential kernel with an automatic relevance determination structure, 'Model A' could extract the highly relevant input features for capacity predictions. Through coupling the Arrhenius law and a polynomial equation into a compositional kernel, 'Model B' is capable of considering the electrochemical and empirical knowledge of battery degradation. The developed models are validated and compared on the Nickel Manganese Cobalt Oxide (NMC) lithium-ion batteries with various cycling patterns. Experimental results demonstrate that the modified Gaussian process regression model considering the battery electrochemical and empirical ageing signature outperforms other counterparts and is able to achieve satisfactory results for both one-step and multi-step predictions. The proposed technique is promising for battery capacity predictions under various cycling cases.

## I. INTRODUCTION

Lithium-ion (Li-ion) batteries have become the primary power sources in electric vehicles (EVs), owing to their advantages such as high energy capability and low self-discharge [1]. However, the lifetime of Li-ion batteries is not unlimited, and their performance is deteriorated along with time and operations [2]. In real applications, a Li-ion battery is generally operated with cyclic charging and discharging operations, further leading its capacity to gradually degrade to the end-of-life (EOL). Without suitable monitor or prognostic solutions for cyclic capacity, both the power and energy of a battery will drop fast [3,4]. Besides, the operational impairment and even catastrophic events are also easy to occur during battery service period [5]. Therefore, to ensure the reliability and safety of a Li-ion battery, reliable and accurate prediction of capacity decay in real-time is critical.

Unfortunately, the battery degradation process is highly nonlinear and influenced by multiple factors, including both manufacturing aspects and operating conditions, which complicate the battery capacity prognostics [6]. Over the last decade, extensive efforts have been carried out to achieve reasonable cyclic capacity predictions, resulting in a myriad of online prognostic approaches. These cyclic capacity prediction techniques can be roughly classified into two categories, namely the model-based and data-driven methods.

For the model-based methods, a proper model with some prior information of a battery, such as the electrochemical models (EMs) [7,8], equivalent circuit models (ECMs) [9,10], together with some recursive observers such as the Kalman filter [11,12] and particle filter [13], are applied to capture and update the battery capacity related parameters based on the data obtained during charging/discharging conditions. Although these model-observer based approaches are able to estimate the cyclic capacity under certain conditions, the model parameters directly determining the estimation performances are still difficult to be accurately identified especially taking into account various cycling conditions. Besides, the involved partial differential equations and large matrix manipulations inevitably increase the computational effort, further restraining their applications in real-time embedded systems for long-term predictions. In addition to the EM/ECM-based approaches, cyclic capacity prediction methods based on the semi-empirical models have also been developed. By means of suitable nonlinear functions (i.e., exponential forms [14,15,16] or polynomial forms [17, 18]) with various data fitting technologies to portray the relation among battery cyclic capacity ageing and stress factors, these referred works seem much easier to be implemented in real-time applications. However, semi-empirical models are inherently open-loop approaches, compromising their generalization and adaptability particularly in long-term predictions.

Data-driven methods, which generally rely on advanced machine learning techniques to capture the underlying mapping among predefined input-output pairs, have also gained

K. Liu is with the Warwick Manufacturing Group, The University of Warwick, Coventry, CV4 7AL, United Kingdom (Email: kliu02@qub.ac.uk, Kailong.Liu@warwick.ac.uk).

increasing attention in the field of battery cyclic capacity prediction. In the literature, various intelligent techniques were successfully devised for cyclic capacity prediction, such as support vector machine (SVM) [19,20] and artificial neural network (ANN) [21,22]. For these state-of-the-art applications, despite the good performance of SVM and ANN on battery capacity prediction, they can only provide deterministic point prediction and failed to provide the corresponding uncertainty quantification. Note that the uncertainties of predicted values could be also helpful for the users to make informed decisions [23]. Therefore, a powerful capacity prediction method calls for not only the prediction of future expectations but also the expression of associated quantified uncertainties.

Deriving from the Bayesian framework, Gaussian process regression (GPR) models have been widely applied to prognostic problems due to their advantages of being nonparametric and probabilistic [24]. The expression of a non-parametric model is naturally adapted to the complexity of data. Therefore, this type of model has the advantage of being more flexible than the parametric ones. The Bayesian approach allows GPR to directly incorporate estimates of uncertainty into predictions, enabling a model to acknowledge the varying probabilities of a range of possible future health values, rather than just giving a single predicted value [25]. Additionally, the structure of GPR is quite simple, as its performance is decided by a mean function and a covariance function. Till now, through constructing specific input features, a limited number of GPR applications exist in the domain of battery management. For instance, in [26], a mixture of GPR model is first presented to study the statistical properties of battery degradation process, followed by using a particle filtering to estimate battery capacities. Through combining the wavelet de-noising approach and GPR, Peng et al. [27] derives a hybrid data-driven approach to predict battery capacities. Based upon the Galvanostatic voltage curve and historical capacity data, Richardson et al. [28,29] propose the conventional covariance function-based GPR models to predict battery cyclic capacities and remaining useful life, respectively. Via constructing four input features from the constant-current constant voltage charging curves, Yang et al. [30] applies the GPR technique to estimate the battery SOH. These publications strongly support the effectiveness of GPR techniques in battery cyclic capacity predictions. However, most studies fit the GPR-based models to the ageing data obtained under similar cyclic conditions, ignoring different cases of stress factors such as temperature and depth-of-discharge (DOD) levels. Such GPR models may not be enough for battery capacity predictions under various cyclic conditions.

Additionally, the existing GPR-based models still fail to exploit many potentials of GPR. For all of the aforementioned GPR technologies, just the conventional kernel functions were studied in battery cyclic capacity predictions. The assumptions of a simple isotropic kernel for the underlying battery degradation behaviours are overly simplified, especially for various cycling cases. To the best of our knowledge, there is still a lack of investigations on combining the capacity-ageing electrochemical or empirical elements with the GPR-based data-driven model to further enhance the accuracy and generalization ability for battery capacity prediction. Hence, benefits of modified covariance functions are insufficiently harnessed in the field of battery cyclic capacity prediction. It should be a promising way by incorporating the battery ageing electrochemical or empirical knowledge into the covariance function to improve the prediction performance of GPR when considering various cyclic conditions.

Considering the foregoing research gap, for the first time, we are strongly incentivized to devise improved GPR-based data-driven methods for Li-ion battery capacity prediction by considering different cyclic conditions (dissimilar temperature and DOD levels). The main contributions of this paper can be summarized as follows. First, to involve the input features of cyclic temperature and DOD, a novel GPR model structure is proposed for predicting battery capacities under various cyclic conditions, where the information of capacity ageing tendency can be also leveraged. Second, two innovative types of GPR-based models are developed based on a systematic understanding of covariance functions: 1) by modifying the basic squared exponential (SE) kernel with the automatic relevance determination (ARD) structure, 'Model A' is capable of removing the irrelevant inputs of GPR; 2) through coupling the Arrhenius law and a polynomial equation with the compositional kernel, 'Model B' is able to integrate the electrochemical and empirical elements of battery ageing into GPR. Finally, the proposed GPR-based techniques are evaluated and compared regarding the prediction accuracy. Obviously, with the coupled electrochemical and empirical elements of battery degradation, 'Model B' outperforms other counterparts in terms of training, one-step, and multi-step prediction results. This is the first known GPR application by constructing its covariance function with the battery ageing electrochemical and empirical knowledge for cyclic capacity prognostics.

The remainder of this article is organized as follows. Section II specifies the cyclic ageing dataset and the proposed GPR model structure. Then the fundamentals behind GPR, two powerful GPR models with improved covariance functions, and the corresponding performance indicators are elaborated in Section III. Section IV provides in-depth analyses of the performance of the proposed GPR-based models for cyclic capacity prediction via two case studies, followed by several comparative studies in Section V. Finally, Section VI concludes this work.

## II. CYCLIC AGEING DATASET AND GPR MODEL STRUCTURE

### A. Cyclic ageing dataset

In real applications, the cycling ageing behaviours of Li-ion batteries are affected by many operational stress factors. In our study, three elements including the operating temperature (°C), DOD (%), and the number of full equivalent cycles (FECs) are considered as the studied stress factors in GPR models to predict capacity degradation dynamics. The cyclic ageing dataset from commercial Nickel-Manganese-Cobalt (NMC)/graphite pouch batteries with a rated nominal capacity of 21Ah is adopted for the purpose of model training and testing. Specifically, eighteen cells were subjected to six different cycling conditions and were grouped as Case 1 to 6. All the cells were cycled with the same middle-SOC of 50%, but under the different cycling DODs (50%, 80% and 100%) and temperatures (35℃ and 45℃). In this manner, all DODs are conducted with an average SOC value of 50%. Detailed cyclic ageing test matrix regarding the cyclic DOD, temperature, charging/discharging current rates are summarized in Table I. Here the capacity degradations were determined by averaging the measured capacities of three cells. The corresponding error bars represent the standard deviations of all three cells in each set, which are within 2% tolerance window and acceptable [14]. It should be known that the effectiveness of this dataset has been proven in [31]. More detailed experimental information can be found in [31], which are not repeated here due to space limitations.

TABLE I
CYCLIC AGEING TEST MATRIX OF TESTED CELLS [31]

| | Cyclic DOD [%] | Temperature [°C] | Charge current rate | Discharge current rate |
|---|---|---|---|---|
| Case 1 | 100 | 35 | C/3 | 1C |
| Case 2 | 50 | 45 | C/3 | 1C |
| Case 3 | 50 | 35 | C/3 | 1C |
| Case 4 | 100 | 45 | C/3 | 1C |
| Case 5 | 80 | 35 | C/3 | 1C |
| Case 6 | 80 | 45 | C/3 | 1C |

On the basis of the above mentioned test matrix, the capacity dataset under six cyclic conditions with different DOD levels (50%, 80% and 100%) and operational temperatures (35℃ and 45℃) is obtained. All these capacity degradation curves, along with the corresponding standard error bars versus cycle number, are shown in Fig. 1. It can be observed that different cyclic conditions would lead to various battery capacity degradation rates. Specifically, for the cyclic conditions of deep cycling DOD levels and high operational temperatures such as Case 4 and Case 6, cells generally present faster degradation trends. In contrast, for the cyclic operations with shallow DOD levels and low temperature such as Case 3 and Case 5, the corresponding capacity degradation rates seem to be relatively slower. According to the comparisons of the capacity degradation curves, battery cyclic capacity degradation process would be accelerated through increasing the operational DOD and temperature levels. For the purpose of exploring ageing information and evaluating the extrapolation performance of the proposed data-driven models, four cases (Case 1, Case 2, Case 3, and Case 4), which are operated under different DOD and temperature levels, are applied as the training dataset to help the model learn the underlying mapping mechanism. The other two cases (Case 5 and Case 6) are used to verify the effectiveness and robustness of the proposed GPR-based model for battery cyclic capacity predictions.

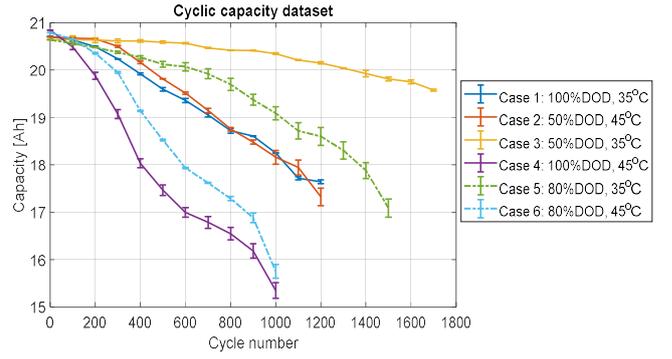

Fig. 1. Cyclic capacity dataset under various DOD and temperature conditions.

### B. GPR model structure

After collecting the cyclic capacity dataset for Li-ion batteries, GPR technologies could be developed to capture the battery capacity dynamics under different cyclic conditions. In order to achieve a reliable and efficient prediction for Li-ion battery capacity degradation, suitable input features require to be extracted in the process of designing the model.

On the one hand, an unsuitable cyclic temperature is detrimental to the battery performance. At high temperature cases, the side-reactions such as transition metal dissolution are accelerated, and the cyclable lithium loss is augmented, further inducing the available capacity loss. Under the low temperature conditions, the dominant capacity degradation mechanism becomes lithium plating, further resulting in the cyclable lithium loss [32]. In light of these, the cyclic temperature is considered as one key input feature of our model.

On the other hand, for many Li-ion battery types, cycling operation would damage the reversibility of electrode active materials, especially at the high DOD level. This could be explained in terms of the mechanical stress induced by the variations occurring in the electrode active materials during battery cyclic operations [33]. Therefore, the cyclic DOD level is also considered as another key input feature.

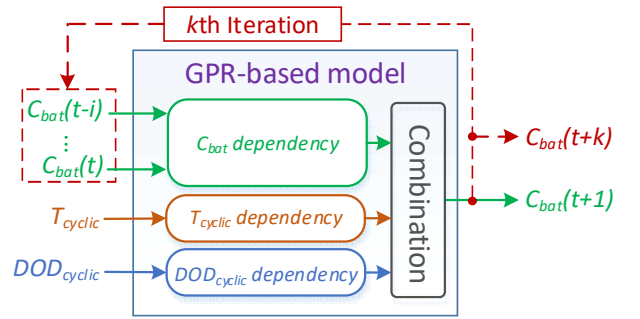

Fig. 2. Model structure for battery cyclic capacity prediction.

Based on the above discussion, an innovative model structure, which not only involves the capacity terms but also takes both cyclic temperature and DOD into account, is proposed for both one-step and multi-step predictions of cyclic capacity degradation, as shown in Fig. 2.

In the training process, after recombination of training dataset (here includes four cases), the corresponding input and output vectors of model are constructed as $[C_{bat}(t-i), ..., C_{bat}(t), T_{cyclic}, DOD_{cyclic}]$ and $C_{bat}(t+1)$, respectively. Here $i$ denotes the length of previous capacity. Both $T_{cyclic}$ and $DOD_{cyclic}$ are constant for each specific cyclic case. Then the GPR is employed to study the underlying mapping among this inputs-output pair offline without adding new data. After training, both one-step $C_{bat}(t+1)$ and multi-step $C_{bat}(t+k)$ predictions are carried out for the testing dataset (here includes two cases). For the multi-step cyclic capacity prediction, a recursive process is carried out using the previously predicted capacity value as the next input point to further predict a new capacity point under the same conditions of $T_{cyclic}$ and $DOD_{cyclic}$. This process will be iteratively conducted until the predefined $k$th step is reached. Then all the further capacity values during this $k$-step predictions could be also attained.

It is evident that in the model structure, previous and current capacity points would be also involved, bringing the benefits that the information of battery capacity degradation tendencies under various $T_{cyclic}$ and $DOD_{cyclic}$ can be simultaneously extracted. Then the GPR-based model would study the potential mechanism among them, giving rising to the fact that the proposed model is able to consider the tendency information under various cyclic conditions.

### III. TECHNOLOGIES

This section first gives a description of the fundamentals behind the GPR technique. Then, two powerful covariance functions are designed within the GPR model for predicting battery capacities under various cyclic conditions. Additionally, the corresponding performance indicators are also elaborated.

*A. Gaussian Process Regression*

Deriving from the Bayesian theory, GPR can be seen as a random process to undertake the nonparametric regression with the Gaussian processes [34]. That is, for any inputs, the corresponding probability distribution over function $f(x)$ follows the Gaussian distribution as:

$$f(x) \sim GPR(m(x), k(x, x')) \quad (1)$$

where $m(x)$ and $k(x, x')$ denote the mean and covariance functions respectively, and expressed by:

$$\begin{cases} m(x) = E(f(x)) \\ k(x, x') = E[(m(x) - f(x'))(m(x) - f(x'))] \end{cases} \quad (2)$$

Here $E()$ represents the expectation value. It is worth noting that in practice, $m(x)$ is generally set to be zero for simplifying calculation process [34]. $k(x, x')$ is also named as the kernel function to explain the relevance degree between a target observation of the training data set and the predicted output based on the similarity of the respective inputs.

In a regression issue, the prior distribution of outputs $y$ can be expressed by:

$$y \sim N(0, k(x, x') + \sigma_n^2 I_n) \quad (3)$$

$N()$ indicates a normal distribution. $\sigma_n$ is the noise term. Supposing there exists a same Gaussian distribution between the testing set $x'$ and training set $x$, the predicted outputs $y'$ would follow a joint prior distribution with the training output $y$ as [30]:

$$\begin{bmatrix} y \\ y' \end{bmatrix} \sim N\left(0, \begin{bmatrix} k(x,x) + \sigma_n^2 I_n & k(x,x') \\ k(x,x')^T & k(x',x') \end{bmatrix}\right) \quad (4)$$

where $k(x,x)$, $k(x',x')$, and $k(x,x')$ represent the covariance matrices among inputs from training set, testing set, as well as training and testing sets, respectively.

In order to guarantee the performance of GPR, some hyperparameters $\theta$ existing in the covariance function require to be optimized by the $n$ points in the training process. One efficient optimization solution is to minimize the negative log marginal likelihood $L(\theta)$ [35] as:

$$\begin{cases} L(\theta) = \frac{1}{2} log[\det \lambda(\theta)] + \frac{1}{2} y^T \lambda^{-1}(\theta) y + \frac{n}{2} log(2\pi) \\ \lambda(\theta) = k(\theta) + \sigma_n^2 I_n \end{cases} \quad (5)$$

After optimizing the hyperparameters of GPR, the predicted output $y'$ can be obtained at dataset $x'$ through calculating the corresponding conditional distribution $p(y'|x', x, y)$ as [30]:

$$p(y'|x', x, y) \sim N(y'|\bar{y}', cov(y')) \quad (6)$$

with

$$\begin{cases} \bar{y}' = k(x, x')^T [k(x,x) + \sigma_n^2 I_n]^{-1} y \\ cov(y') = k(x', x') - k(x, x')^T [k(x,x) + \sigma_n^2 I_n]^{-1} k(x, x') \end{cases}$$

where $\bar{y}'$ stands for the corresponding mean values of prediction. $cov(y')$ denotes a variance matrix to reflect the uncertainty range of these predictions. More details on these equations of GPR can be found in [34].

*B. Model A' with ARD-SE kernel*

Noted that the covariance function has to be selected or designed carefully as it plays a vital role in determining the performance of GPR. In fact, there are several common covariance functions in the GPR, but the selection of a suitable one is a case-by-case issue.

One efficient common covariance function is the squared exponential (SE) kernel with a radial basis form [34] as:

$$k_{SE}(x, x') = \sigma_f^2 exp\left(-\frac{\|x - x'\|^2}{2\sigma_l^2}\right) \quad (7)$$

where $\sigma_f$ is a hyperparameter to control the amplitude of covariance for all kernel case. $\sigma_l$ is another hyperparameter to reflect the spread.

In practice, although SE kernel is smooth with simple structure, the basic SE kernel is still lack of ability to provide reliable results especially for the highly nonlinear mapping that involves multi-dimensional input terms. This is mainly caused by the limited expressiveness of the single isotropic SE function. In our cyclic ageing prediction domain, input variables are not only composed of the time-series battery capacity values, but also involving the corresponding operational temperatures and DODs. To extract different features and improve the prediction performance, one efficient way is to modify the basic SE kernel with the automatic relevance determination (ARD) structure [36], as expressed by:

$$k_{ARD\sim SE}(x, x') = \sigma_f^2 \exp\left(-\frac{1}{2}\sum_{d=1}^{D}\frac{\|x_d - x'_d\|^2}{\sigma_d^2}\right) \quad (8)$$

where $D$ denotes the number of input terms. It is evident that each input term $x_d$ has an hyperparameter $\sigma_d$ corresponding to that input term. Generally, $\sigma_d$ is capable of determining the relevancies of input term to the regression results. A small $\sigma_d$ results in a high relevancy.

Due to the in-built benefits of determining the relevancies for multi-dimensional input variables, ARD-SE kernel has become a competitive covariance function. Therefore, for battery cyclic ageing prediction, ARD-SE kernel may be a good candidate to extract features and capture ageing dynamics under various cyclic conditions.

In light of this, an attempt has been made first in this study to establish a GPR-based model by using the ARD-SE kernel for battery cyclic capacity prediction as:

$$k_{ARD\sim SE}(\boldsymbol{x}, \boldsymbol{x'}) = \sigma_f^2 \exp\left[-\frac{1}{2}\left(\frac{\|x_T - x'_T\|^2}{\sigma_T^2} + \frac{\|x_{DOD} - x'_{DOD}\|^2}{\sigma_{DOD}^2} + \sum_{c=1}^{i+1}\frac{\|x_C - x'_C\|^2}{\sigma_C^2}\right)\right] \quad (9)$$

where $\sigma_f, \sigma_T, \sigma_{DOD}$ and $\sigma_C$ denote the corresponding hyperparameters. $i+1$ stands for the number of input capacity terms with the same form as that in Fig. 2. Here we name this type of GPR model as 'Model A'. Theoretically, through using 'Model A' in cyclic capacity prediction domain, irrelevant features among battery capacity, operational temperature and DOD could be restrained by setting large lengthscales for them, resulting in a relatively sparse and explanatory subset of features. Additionally, the prediction accuracy and generalization of 'Model A' can be improved to some extent because different predictors with various lengthscales are generated by using ARD structure.

### C. 'Model B' with modified kernel

As described in Subsection III.B, we formulate a competitive GPR model (Model A) by using the ARD-SE kernel. However, for most literature, the kernel functions in GPR were usually selected through using the trial and error approaches, without considering any battery physical or chemical mechanisms of the related applications. Following this way, it is difficult to guarantee the applicability of the selected kernel in the general context of Li-ion battery cyclic capacity prediction. To improve the performance of GPR-based model in cyclic capacity prediction domain, a suitable modified kernel considering the electrochemical and empirical information of Li-ion battery degradation is necessary and encouraged.

On the basis of our proposed model structure in Subsection II.B, apart from time series capacities $[C_{bat}(t-i), \ldots, C_{bat}(t)]$ as the inputs to tell model the information of degradation trend, there also exists other two input terms including the $T_{cyclic}$ and the $DOD_{cyclic}$ to contain the information of operational temperature and DOD. These two stress factors are essential in determining the battery cyclic capacity dynamics. Therefore, the effects of both these factors are required to be carefully considered. It should be known that the electrochemical or empirical knowledge corresponding to these input terms is able to benefit the constructions of suitable kernel function within the GPR model. To this end, an attempt has been made in this study to modify the kernel function within GPR to derive an innovative model (labelled as 'Model B') that considers the electrochemical or empirical elements of Li-ion battery ageing. Specifically, the related components within a kernel function are modified separately to reflect all input terms including the temperature, DOD, and battery capacities.

**Temperature dependency:** during cycling process, temperature is a key factor to affect battery dynamics. To effectively consider this dependency, the Arrhenius law $f_A(T)$, which illustrates that the side reactions within battery would exponentially reduce with decreasing temperature, has been reported in a myriad of literatures such as [14,17] to describe the effects of temperature as:

$$f_A(T) = a \cdot \exp(-E_A/RT) \quad (10)$$

where $a$ is a pre-exponential factor. $R$ stands for the ideal gas constant. $E_A$ denotes the activation energy of an electrochemical reaction. $T$ is the operational temperature condition.

On the basis of this Arrhenius equation, the component $k_{Tcyc}(x_T, x'_T)$ related to $T_{cyclic}$ in the kernel function would be modified with the same exponential form as:

$$k_{Tcyc}(x_T, x'_T) = l_T \cdot \exp\left(-\frac{1}{\sigma_T}\left\|\frac{1}{x_T} - \frac{1}{x'_T}\right\|\right) \quad (11)$$

where $l_T$ and $\sigma_T$ are corresponding hyperparameters. It is worth noting that this component of temperature dependency is expressed with an isotropic form to explain the relevance degree between outputs based on the difference between their respective cyclic temperature conditions $x_T$ and $x'_T$. Both these cyclic temperatures are first transferred to the reciprocal forms before introducing to a common Laplacian kernel. In the light of this, the Arrhenius law has been successfully coupled within the GPR model.

**DOD dependency:** next, we try to construct a suitable component within GPR to take DOD dependency into account. DOD is another essential point to battery cyclic ageing. For most battery types, a large DOD would result in a disproportional stronger ageing in comparison with the small DOD. According to many existing reports [14,37], the effect of DOD on battery cyclic ageing generally presents a polynomial or linear tendency. That is, the empirical knowledge of operational DOD dependency can be well described by using a polynomial equation. Therefore, to consider the DOD dependency in the covariance function of GPR, a polynomial kernel is applied in our study. Then a specific component $k_{DOD}(x_{DOD}, x'_{DOD})$ related to $DOD_{cyclic}$ is modified with the polynomial form as:

$$k_{DOD}(x_{DOD}, x'_{DOD}) = (l_D \cdot x_{DO}^T \, x'_{DOD} + c_D)^{d_D} \quad (12)$$

where the slope $l_D$, the constant term $c_D$ and the polynomial degree $d_D$ are the hyperparameters of this component. It should be known that this component belongs to a non-stationary kernel, bringing the benefits that the computation effort can be reduced because small quantity of training data is required for non-stationary kernel [38].

**Capacity dependency:** For our proposed model structure,

one key benefit by using previous and current capacities as input terms is that the corresponding ageing trend information can be studied by GPR model in the training stage. In order to reflect this correlation, the components within covariance function for these capacity terms $x_c$ and $x'_c$ are all described by the SE kernel with hyperparameters $l_c, \sigma_c$ as:

$$k_{Cbat}(x_c, x'_c) = l_c^2 \, exp\left(-\sum_{c=1}^{i+1} \frac{\|x_c - x'_c\|^2}{2\sigma_c^2}\right) \quad (13)$$

At this point, the component of kernel function within GPR model for each input term has been formulated through the electrochemical or empirical knowledge of Li-ion battery degradation. After that, it is vital to combine these components to derive a suitable kernel function. According to the guidance in [38], the kernel function of GPR has to be positive semi-definite. Adding or multiplying the basic kernels is an effective way to result in a positive semi-definite compositional kernel. On the one hand, adding basic kernels could be seen as an 'Or' operation, which means that these summed basic kernels are independent with each other. On the other hand, multiplying basic kernels is similar as the 'And' operation, bringing the benefits that the correlations among each input terms can be simultaneously taken into account.

In battery cyclic ageing prediction, it is obvious that there exists the strong underlying correlations among cyclic temperature, DOD and battery capacities. In the light of this, multiplying each component to generate a compositional kernel is suggested in this study to consider the interactions among these input terms. Then a novel compositional kernel for 'Model B' is formulated with the specific form as:

$$k_{modified}(\pmb{x}, \pmb{x'}) = k_{Cbat}(x_c, x'_c) \cdot k_{Tcyc}(x_T, x'_T)$$
$$\cdot k_{DOD}(x_{DOD}, x'_{DOD})$$
$$= l_f^2 \cdot exp\left(-\frac{1}{\sigma_T}\left\|\frac{1}{x_T} - \frac{1}{x'_T}\right\|\right) \cdot (x_{DOD}^T x'_{DOD} + c_D)^{d_D} \cdot$$
$$exp\left(-\sum_{c=1}^{i+1} \frac{\|x_c - x'_c\|^2}{\sigma_c^2}\right) \quad (14)$$

where $\pmb{x}$ and $\pmb{x'}$ denote the input vectors with the same terms as $\pmb{x} = (x_c, \; x_T, \; x_{DOD})$. Under the proposed model structure as shown in Fig. 2, $x_c$ is $[C_{bat}(t-i), \dots, C_{bat}(t)]$, $x_T = T_{cyclic}$, and $x_{DOD} = DOD_{cyclic}$, respectively. $l_f, \sigma_T, c_D, d_D$ and $\sigma_C$ are the corresponding hyperparameters. Obviously, 'Model B' can be used to predict battery cyclic capacity with the considerations of Li-ion battery electrochemical or empirical ageing elements in our study.

### D. Performance indicators

In order to evaluate the results of our proposed GPR models, the predicted capacity values should be compared with the corresponding real data from the experiment. In this regard, three different performance metrics are applied in this study for the evaluation purpose [31].

1) Mean absolute error (MAE): by defining as (15), the MAE indicator is generally used to reflect the actual situations of predicted errors. The larger the MAE, the less accurate the predicted results.

$$MAE = \frac{1}{M}\sum_{i=1}^{M}|y_i - \hat{y}_i| \quad (15)$$

where $M$ denotes the total number of predictions. $y_i$ stands for the actual test values, and $\hat{y}_i$ represents the predicted values from GPR model. These definitions are also suitable for the following indicators.

2) Maximum absolute error (ME): by defining as (16), ME is used to reflect the maximum difference between predicted values and real values. The larger the ME, the bigger difference occurs during prediction.

$$ME = \max_{1 \leq i \leq M}|y_i - \hat{y}_i| \quad (16)$$

3) Root mean square error (RMSE): by defining as (17), RMSE is another popular indicator to reflect the deviation between the predicted values and real values. In comparison with MAE, larger absolute values would be penalized through giving more weight by using RMSE.

$$RMSE = \sqrt{\frac{1}{M}\sum_{i=1}^{M}(y_i - \hat{y}_i)^2} \quad (17)$$

## IV. PREDICTION RESULTS AND ANALYSIS

As discussed in the previous section, the GPR model with the common kernel is more or less restricted, thus may not be the most proper solutions for capacity predictions of Li-ion battery under different cyclic conditions. In contrast, the GPR model through using the improved kernel is significantly appealing due to the ability to consider the electrochemical or empirical knowledge of battery cyclic ageing. This section in-depth investigates the performance of GPR-based model for cyclic capacity prediction via two case studies, with the special focuses on the model's accuracy.

### A. Case study 1 - Model A

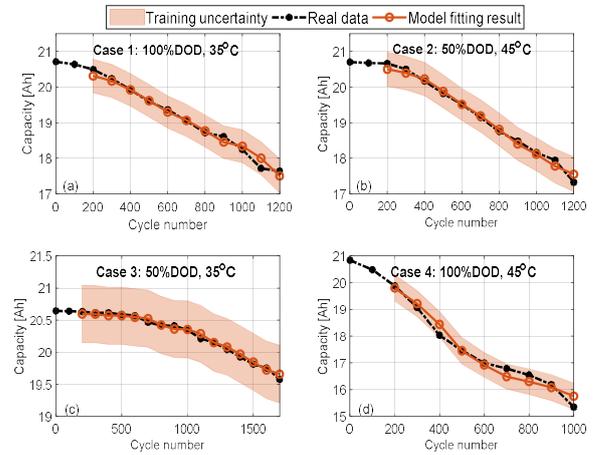

Fig. 3. Model structure for battery cyclic capacity prediction.

TABLE II
PERFORMANCE INDICATORS FOR TRAINING RESULTS BY USING 'MODEL A'

| Training cases | Case 1 | Case 2 | Case 3 | Case 4 |
|---|---|---|---|---|
| **ME [Ah]** | 0.2847 | 0.2265 | 0.0808 | 0.3053 |
| **MAE [Ah]** | 0.0870 | 0.0928 | 0.0357 | 0.0966 |
| **RMSE [Ah]** | 0.1270 | 0.1121 | 0.0420 | 0.1409 |

The first case study aims to evaluate the performance of GPR-based 'Model A' that is constructed with the ARD-SE kernel. Through using the ARD structure, 'Model A' owns a strong feature extraction ability in theory. Fig. 3 and Table II show the training results and performance indicators of 'Model

A' for different cyclic temperature and DOD cases. Through using the ARD-SE kernel, 'Model A' is capable of capturing the overall capacity ageing trends. Here the maximum difference between the upper and lower confidence bounds almost reaches 1 Ah for all these cases. The uncertainty boundaries of training cases are mainly related to the 'scope compliance' uncertainty, which is used to quantify "how confident" the GPR model felt when performing predictions [23]. According to the corresponding performance indicators in Table II, Case 3 achieves the best training results. In contrast, Case 4 owns the largest ME of 0.3053Ah, which is obtained around the 400th cycle. The MAE and RMSE values for Case 4 also reach the maximum ones with 0.0966Ah and 0.1409Ah respectively, which are almost 11.2% and 10.9% larger than those from Case 1. However, due to the different lengthscale is assigned to each component within the ARD structure, all the RMSE values for such training cases are still less than 0.15 Ah (0.71% of total capacity), indicating a reliable overall training result with reasonable accuracy. Therefore, the 'Model A' with ARD-SE kernel is effective for battery cyclic capacity training.

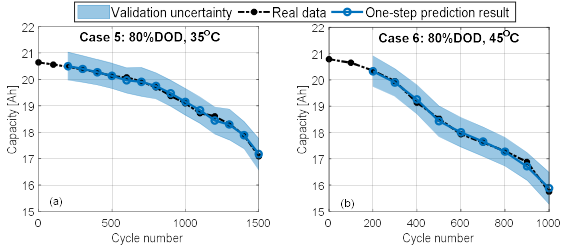

Fig. 4. One-step prediction results by using 'Model A' for testing dataset.

After training process, 'Model A' is applied to predict the future capacities for testing dataset. To investigate the extrapolation performance of ARD-SE kernel, both the one-step and multi-step predictions are carried out with the detailed analyses of their prediction results for two specific cases including Case 5 and Case 6. Table III illustrates the performance indicators for both one-step and multi-step predictions of 'Model A'. The one-step prediction results for this testing dataset by using the well-trained 'Model A' are illustrated in Fig. 4. In this study, battery capacity values for all cyclic cases are between 15 Ah and 21Ah without significant difference. Besides, five-fold cross-validation is adopted to avoid training over-fitting. Therefore, there exists no significant uncertainty interval difference between the training and one-step prediction cases. It is observed that the cyclic capacity degradation trend for testing dataset can be well tracked by ARD-SE kernel. Specifically, ME values in one-step predictions for Case 5 and Case 6 are 0.2410Ah and 0.1689Ah, respectively. In addition, the RMSEs for these two cases are both less than 0.1Ah (0.5% of total capacity), indicating that the one-step predictions by using the proposed 'Model A' is reliable.

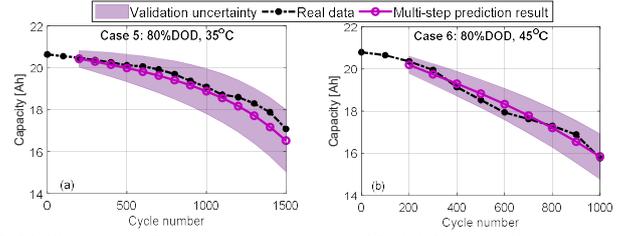

Fig. 5. Multi-step prediction results by using 'Model A' for testing dataset.

TABLE III
PERFORMANCE INDICATORS FOR ONE-STEP AND MULTI-STEP PREDICTION RESULTS BY USING 'MODEL A'

| Testing cases | Case 5 | | Case 6 | |
|---|---|---|---|---|
| Prediction types | One-step | Multi-step | One-step | Multi-step |
| ME [Ah] | 0.2410 | 0.7073 | 0.1689 | 0.3919 |
| MAE [Ah] | 0.0667 | 0.2996 | 0.0794 | 0.2115 |
| RMSE [Ah] | 0.0974 | 0.3618 | 0.0936 | 0.2350 |

Next, we investigate the effectiveness of 'Model A' for multi-step capacity predictions. For the iterative multi-step prediction, the new estimation will be utilized for the next step prediction. Thus such uncertainty is accumulated in each step. Considering this covariance propagation behaviour would result in the uncertainty bounds become larger as the prediction horizon increases. In this test, after putting two capacity terms along with the corresponding $T_{cyclic}$ and $DOD_{cyclic}$ into the well-trained GPR model, the predictions will be conducted iteratively to obtain all future battery cyclic capacities. In order to verify model's performance based on the real collected dataset, final step numbers of Case 5 and Case 6 in multi-step predictions are set as 14 and 9, respectively. It should be known that with the same iterative way as many related publications [21], this algorithm can actually continue to predict the lifetime of a battery for a specific application. Besides, to consider the covariance propagation behaviour in iterative multi-step predictions, an effective method in [39] through adding a propagation term which depends on the derivatives of covariance function is adopted, thus the uncertainty could be propagated and continuously accumulated. More information and detailed analysis of the uncertainty propagation can be found in the Eq. (11) of Section IV and the Appendix of our group's work [39]. Readers who have interests could check [39] and the references therein. Fig. 5 presents the multi-step prediction results for 'Case 5' and 'Case 6'. It is clear that for Case 5, 'Model A' with the ARD-SE kernel cannot well capture the true capacity points all the time. Specifically, the prediction results are acceptable before 1100 cycles but after that, large mismatches occur. Quantitatively, the ME, MAE and RMSE are 193.5%, 349.2% and 271.5% larger than those from one-step prediction case. From Fig. 5(b), although the global capacity degradation trend of Case 6 is captured by 'Model A', there still exists large mismatch points. Here the ME, MAE and RMSE are 0.3919Ah, 0.2115Ah and 0.2350Ah, which are still 132.0%, 166.4% and 151.1% larger than the one-step prediction. Accordingly, 'Model A' with the ARD-SE kernel fails to capture the prior information of global degradation trend, further decreasing its performance for multi-step predictions.

## B. Case study 2 - Model B

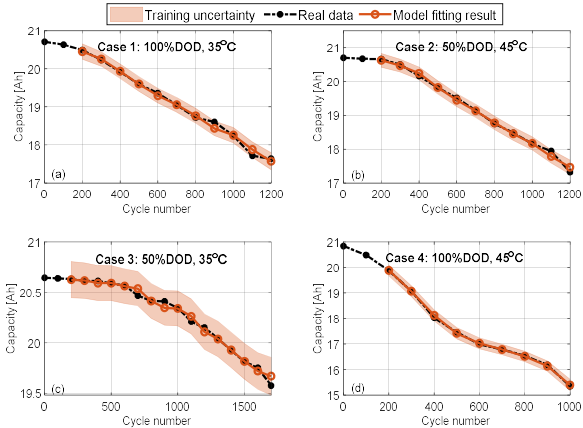

Fig. 6. Training results by using 'Model B' for each cyclic cases from training dataset.

TABLE IV
PERFORMANCE INDICATORS FOR TRAINING RESULTS BY USING 'MODEL B'

| Training cases | Case 1 | Case 2 | Case 3 | Case 4 |
|---|---|---|---|---|
| ME [Ah] | 0.1689 | 0.1575 | 0.0928 | 0.0915 |
| MAE [Ah] | 0.0557 | 0.0531 | 0.0259 | 0.0347 |
| RMSE [Ah] | 0.0790 | 0.0731 | 0.0386 | 0.0442 |

The performance of 'Model B' is also evaluated for the same training and testing dataset. Fig. 6 and Table IV illustrate its training results and the corresponding performance indicators. In general, through using the improved kernel to consider the electrochemical and empirical knowledge of battery cyclic ageing, 'Model B' is able to describe the capacity fading dynamics better than 'Model A'. Such superiority is apparent with more accurate fitting results. Quantitatively, the maximum ME, MAE and RMSE of all training cases by using 'Model B' are obtained for Case 1 with just 0.1689Ah, 0.0557Ah and 0.0790Ah, which are 40.7%, 36.0% and 37.8% less than those from 'Model A'.

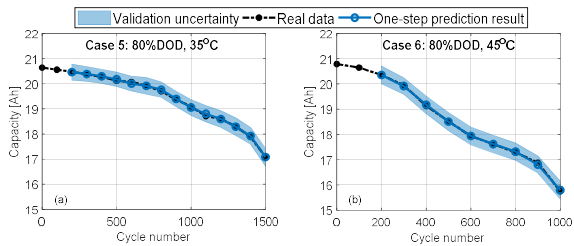

Fig. 7. One-step prediction results by using 'Model B' for testing dataset.

TABLE V
PERFORMANCE INDICATORS FOR ONE-STEP AND MULTI-STEP PREDICTION RESULTS BY USING 'MODEL B'

| Testing cases | Case 5 | | Case 6 | |
|---|---|---|---|---|
| Prediction types | One-step | Multi-step | One-step | Multi-step |
| ME [Ah] | 0.1475 | 0.2576 | 0.0751 | 0.2004 |
| MAE [Ah] | 0.0447 | 0.0680 | 0.0273 | 0.0512 |
| RMSE [Ah] | 0.0598 | 0.0873 | 0.0355 | 0.0771 |

After training, the similar one-step and multi-step prediction tests are conducted to investigate the extrapolation performance of 'Model B'. Fig. 7 shows the results by using 'Model B' for one-step cyclic capacity prediction. It is evident that for both Case 5 and Case 6, the trained 'Model B' is able to well capture the evolution of cyclic capacity ageing trends, as indicated by the effective match between the predicted capacity points and the real data. The corresponding performance indicators of both one-step and multi-step predictions are presented in Table V. Not surprisingly, the ME, MAE and RMSE for all one-step prediction cases are within 0.15Ah, 0.05Ah and 0.06Ah, respectively, which are 11.2%, 25.0% and 28.4% less than those from 'Model A' case. Therefore, 'Model B' is highly accurate for the one-step cyclic capacity prediction.

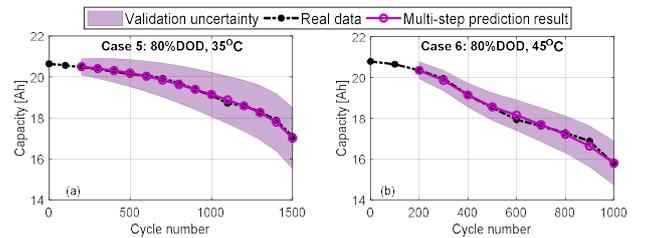

Fig. 8. Multi-step prediction results by using 'Model B' for testing dataset.

Next, we focus on the evaluation of multi-step prediction performance for 'Model B'. For this test, the same multi-step prediction process that mentioned in the above subsection is iteratively conducted by using the improved kernel. As illustrated in Fig. 8, the predicted capacities become highly similar to the real data for both Case 5 and Case 6, indicating the satisfactory multi-prediction accuracy is obtained by using our proposed 'Model B'. From Fig. 8(a), apart from some small mismatches occur at the large local fluctuations, the global prediction performance is highly improved in comparison with that from 'Model A'. Here the MAE and RMSE for Case 5 are 0.0680Ah and 0.0873Ah, which are 77.3% and 75.9% less than those from multi-step prediction case of 'Model A'. Similarly, more effective multi-step prediction results are also observed in Case 6. Quantitatively, 'Model B' achieves the ME, MAE and RMSE values of 0.2004Ah, 0.0512Ah and 0.0771Ah for the Case 6 test. In comparison with the results of 'Model A' using the ARD-SE kernel, 'Model B' offers at least 46.9% increase in accuracy.

## V. COMPARISON ANALYSIS AND DISCUSSION

In this section, to quantify the effects and determine the suitable number of input capacity terms, a comparative study among different number of input terms is first conducted. Then, to further illustrate the effectiveness of modified GPR-based models for battery cyclic capacity predictions, a conventional GPR model with the single SE kernel is also adopted and compared under the same dataset.

### A. Comparative study with different number of input capacity terms

For our proposed model structure as illustrated in Fig. 2, the number of previous and current capacity terms (here equates to the value of $i + 1$ in Fig. 2.) plays a vital role in determining the capacity prediction performance. Generally, on the one hand, increasing the number of input capacity terms would bring more information regarding the capacity degradation tendency in the learning process, further enhancing the accuracy of predicted results. On the other hand, a data-driven model with lots of input terms would increase the model complexity and computational effort. Besides, too much input terms may also result in the overfitting problem, further reducing the model's performance. In the light of this, it is vital to select a suitable number of input capacity terms for our proposed model.

To evaluate the effects of the different number of input terms and further prevent model from overfitting, five values of input capacity terms starting from one ($i = 0$) to five ($i = 4$) are selected. Here we take the multi-step prediction results of 'Model B' as illustration because the prediction performance of this electrochemical mechanism-involved GPR model is of the utmost concern in this study. After training 'Model B' based on the constructed inputs-output pairs, a multi-step prediction process similar to that described in Subsection IV.B is conducted for each case of input number.

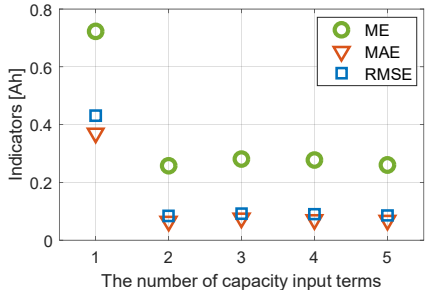

Fig. 9. Indicators of using 'Model B' with different input capacity terms for multi-step predictions.

All three indicators (ME, MAE and RMSE) for the prediction results by using 'Model B' with various input capacity terms are plotted in Fig. 9. Interestingly, it can be observed that using two capacity input terms would significantly decrease the values of all indicators in comparison with using only one capacity term. After that, further adding capacity terms as inputs would not cause large variations of these indicators. Conversely, all these indicators, more or less, present slight increase tendencies. That is, involving more than two capacity input terms may cause overfitting problem, further decreasing the prediction performance of our proposed model. Therefore, the number of input capacity terms is set as two in this study to not only guarantee the predicted accuracy but also restrain overfitting.

*B. Comparison with SE-based GPR model*

In order to further evaluate the performance of improved GPR models, a conventional single SE-based GPR model (SEGM) with two hyperparameters ($\sigma_f$ and $\sigma_l$) in the form of (18) is also applied based on the same dataset. The comparisons in terms of both training and prediction results would be conducted among the SEGM, 'Model A' and 'Model B' in this study. Table VI illustrates the corresponding hyperparameters within the kernels of these GPR-based models.

$$k_{SE}(\boldsymbol{x}, \boldsymbol{x}') = \sigma_f^2 \exp\left(-\frac{\sum_{c=1}^{i+1}\|x_c - x'_c\|^2 + \|x_T - x'_T\|^2 + \|x_{DOD} - x'_{DOD}\|^2}{2\sigma_l^2}\right)$$

(18)

TABLE VI
HYPERPARAMETERS OF GPR-BASED MODELS

| Model types | Hyper-parameters |
|---|---|
| **SEGM** | $\sigma_f = 0.894, \sigma_l = 2.036$ |
| **Model A** | $\sigma_f = 0.826, \sigma_T = 1.813, \sigma_{DOD} = 2.391,$ $\sigma_1 = 2.489, \sigma_2 = 1.430$ |
| **Model B** | $l_f = 0.516, \sigma_T = 3.964, c_D = 4.520, d_D = 1.323,$ $\sigma_1 = 5.282, \sigma_2 = 3.351$ |

*1) Comparison with the Training Results:* To evaluate and compare the training performance of each model type, the indicators (ME, MAE and RMSE) for SEGM, 'Model A' and 'Model B' in terms of the training results of total dataset are shown in Fig. 10. Obviously, through using the improved GPR technique to learn the underlying mapping among capacity, DOD and temperature, the training results of 'Model A' and 'Model B' are both better than those from SEGM. Quantitatively, the ME, MAE and RMSE for total training dataset by using SEGM are 0.6481Ah, 0.2290Ah and 0.2712Ah, respectively. Through using GPR models with the improved kernels, it turns out that the corresponding training results improve nearly twice as good as the SEGM. Furthermore, in comparison with the ME, MAE and RMSE values of 0.4053Ah, 0.1061Ah and 0.1313Ah for the 'Model A', 'Model B' has the ME, MAE and RMSE values of 0.1689Ah (58.3% decrease), 0.0428Ah (59.7% decrease) and 0.0589Ah (55.1% decrease) for the total training dataset. The two modified GPR models hence validate themselves with higher training capability.

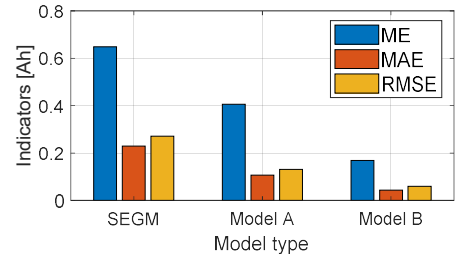

Fig. 10. Indicators of using different model types for all training dataset.

*2) Comparison with the Prediction Results:* For the sake of intuitively evaluating the multi-step prediction performance of each model type, the corresponding performance indicators for total testing dataset are compared and shown in Fig. 11. Due to the participation of predicted data in multi-step prediction cases, it is well expected that the ME, MAE and RMSE are all larger than those from training results. By using the improved kernels, the multi-step prediction performances of both 'Model A' and 'Model B' are enhanced as desired. From Fig. 11, the ME and RMSE for these two GPR models are within 0.71Ah and 0.31Ah, which are 32.6% and 13.6% less than those from

SEGM. Moreover, comparing to the 'Model A' with ARD-SE kernel, 'Model B' also achieves the significant improvements for multi-step cyclic capacity prediction cases. In this case, the ME, MAE and RMSE values become 0.2576Ah (63.6% decrease), 0.0652Ah (74.7% decrease) and 0.0835Ah (72.3% decrease) respectively, which validates the superiority of coupling the electrochemical or empirical elements into the covariance function of GPR.

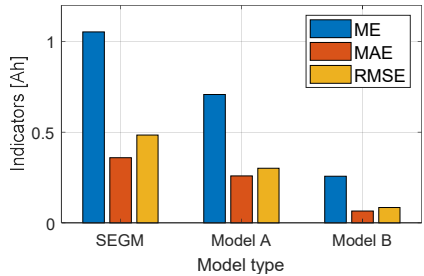

Fig. 11. Indicators of using different model types for multi-step prediction.

On the basis of above discussions, both 'Model A' and 'Model B' outperform SEGM for the same training and testing datasets. This suggests that the GPR model with the modified kernel is promising for battery capacity predictions under various cyclic cases.

*C. Further Discussions*

This article focuses on the development of modified kernel functions of GPR to achieve reliable battery cyclic capacity predictions under various temperature and DOD conditions. For the first time, electrochemical and empirical elements of battery ageing are effectively coupled into the covariance functions of GPR. Through using the suitable GPR solvers such as the Matlab GPR toolbox with a 2.40 GHz Intel Pentium 4 CPU, all components can be effectively integrated into the compositional kernel and the GPR models can be well trained within 15s. Noting that the obvious battery cyclic ageing generally takes several days or months, while the computational effort of GPR-based model is only about a few seconds. In this case, the proposed GPR-based capacity estimator can be conveniently implemented in real time application. In this study, due to the limited experimental resources, we mainly focus on the stress factors of DOD and relatively higher temperature. This makes our proposed models do not have a specific requirement on the battery charge/discharge currents. Besides, none of the modifications can be seen as a one-size-fits-all solution for GPRs. How to effectively improve GPR performance is still an open research question for battery capacity prognostics.

More further interesting areas for the GPR-based battery capacity predictions can be summarized as: 1) Even though testing the battery ageing performance with more stress factor cases is time-consuming, extensive ageing tests are also recommended to generate more well-rounded database especially for real EV applications. 2) After collecting valuable ageing data of EV applications, more useful input features such as the absolute elapsed time during certain EV conditions, charge throughput, mileage or cycle number etc. could be extracted for battery capacity prognostics. 3) As mean function is able to dominate the predictions in regions far from the training data, developing suitable $m(x)$ for GPR can be considered especially in long-term battery ageing prediction domain. Based on the above solutions, our future work includes the improvement of GPRs to consider more stress factors such as charging/discharging current rate, lower temperatures and even factors changing cases. Besides, inspired by the process in [39], an in-depth systematic analysis on the uncertainty propagation of iterative multi-step battery capacity predictions would be conducted in the future.

## VI. CONCLUDING REMARKS

In this article, the novel data-driven approaches are presented for predicting the cyclic capacity of the NMC-based Li-ion batteries under various operational temperature and DOD conditions. The modified GPR models are employed to study the underlying relationship among degraded battery capacity, cyclic temperatures and DODs. In particular, on the basis of electrochemical and empirical knowledge of battery degradation, an attempt has been made to construct a mechanism-conscious GPR model in cyclic capacity prediction domain. Through coupling the Arrhenius law and polynomial equation into a compositional kernel within GPR, 'Model B' allows to make satisfactory predictions and reliable uncertainty quantification of capacity degradation for the Li-ion batteries subjected to various cyclic conditions. Illustrative results show that the proposed data-driven models are effective and robust. It is also confirmed that the modified covariance function considering the battery mechanism knowledge outperforms the ARD-SE based covariance function in both training and testing phases. Here the ME, MAE, RMSE of 'Model B' are less than 0.26Ah (1.2%), 0.07Ah (0.3%) and 0.09Ah (0.4%) for all multi-step prediction cases. This is the first known application by directly coupling the battery electrochemical and empirical cycling ageing information into the GPR. After collecting suitable battery ageing data, battery manufacturers are able to use our proposed GPR models to their operational conditions and achieve satisfactory predictions as well as reliable uncertainty management regarding the future cyclic capacities and battery SOH.

## REFERENCE

[1] X. Hu, C. Zou, C. Zhang, and Y. Li. "Technological developments in batteries: a survey of principal roles, types, and management needs." IEEE Power and Energy Magazine 15, no. 5 (2017): 20-31.
[2] Z. Wei, J. Zhao, R. Xiong, G. Dong, J. Pou, and K. J. Tseng. "Online estimation of power capacity with noise effect attenuation for lithium-ion battery." IEEE Transactions on Industrial Electronics 66, no. 7 (2018): 5724-5735.
[3] Z. Yang, D. Patil, and B. Fahimi. "Online estimation of capacity fade and power fade of lithium-ion batteries based on input–output response technique." IEEE Transactions on Transportation Electrification 4, no. 1 (2017): 147-156.
[4] K. Liu, C. Zou, K. Li, and T. Wik. "Charging pattern optimization for lithium-ion batteries with an electrothermal-aging model." IEEE Transactions on Industrial Informatics 14, no. 12 (2018): 5463-5474.

[5] A. Barré, B. Deguilhem, S. Grolleau, M. Gérard, F. Suard, and D. Riu. "A review on lithium-ion battery ageing mechanisms and estimations for automotive applications." Journal of Power Sources 241 (2013): 680-689.
[6] M. Jafari, A. Gauchia, S. Zhao, K. Zhang, and L. Gauchia. "Electric vehicle battery cycle aging evaluation in real-world daily driving and vehicle-to-grid services." IEEE Transactions on Transportation Electrification 4, no. 1 (2017): 122-134.
[7] R. Gu, P. Malysz, H. Yang, and A. Emadi. "On the suitability of electrochemical-based modeling for lithium-ion batteries." IEEE Transactions on Transportation Electrification 2, no. 4 (2016): 417-431.
[8] R. Xiong, L. Li, Z. Li, Q. Yu, and H. Mu. "An electrochemical model based degradation state identification method of Lithium-ion battery for all-climate electric vehicles application." Applied energy 219 (2018): 264-275.
[9] M. Bahramipanah, D. Torregrossa, R. Cherkaoui, and M. Paolone. "Enhanced equivalent electrical circuit model of lithium-based batteries accounting for charge redistribution, state-of-health, and temperature effects." IEEE Transactions on Transportation Electrification 3, no. 3 (2017): 589-599.
[10] X. Hu, F. Fei, K. Liu, L. Zhang, J. Xie, and B. Liu. "State estimation for advanced battery management: Key challenges and future trends." Renewable and Sustainable Energy Reviews 114 (2019): 109334.
[11] Y. Chang, H. Fang, and Y. Zhang. "A new hybrid method for the prediction of the remaining useful life of a lithium-ion battery." Applied Energy 206 (2017): 1564-1578.
[12] K. Liu, K. Li, Q. Peng, and C. Zhang. "A brief review on key technologies in the battery management system of electric vehicles." Frontiers of Mechanical Engineering 14, no. 1 (2019): 47-64.
[13] X. Tang, C. Zou, K. Yao, J. Lu, Y. Xia, and F. Gao, "Aging trajectory prediction for lithium-ion batteries via model migration and Bayesian Monte Carlo method," Applied Energy, 254（2019）: 113591.
[14] J. Schmalstieg, S. Käbitz, M. Ecker, D. U. Sauer. "A holistic aging model for Li (NiMnCo) O2 based 18650 lithium-ion batteries." Journal of Power Sources 257 (2014): 325-334.
[15] A. Guha, and A. Patra. "State of health estimation of lithium-ion batteries using capacity fade and internal resistance growth models." IEEE Transactions on Transportation Electrification 4, no. 1 (2017): 135-146.
[16] K. Liu, X. Hu, Z. Yang, Y. Xie, and S. Feng. "Lithium-ion battery charging management considering economic costs of electrical energy loss and battery degradation." Energy Conversion and Management 195 (2019): 167-179.
[17] M. Ecker, J. B. Gerschler, J. Vogel, S. Käbitz, F. Hust, P. Dechent, D. U. Sauer. "Development of a lifetime prediction model for lithium-ion batteries based on extended accelerated aging test data." Journal of Power Sources 215 (2012): 248-257.
[18] J. de Hoog, J. M. Timmermans, D. Ioan-Stroe, M. Swierczynski, J. Jaguemont, S. Goutam, N. Omar, J. V. Mierlo, and P. V. D. Bossche. "Combined cycling and calendar capacity fade modeling of a Nickel-Manganese-Cobalt Oxide Cell with real-life profile validation." Applied Energy 200 (2017): 47-61.
[19] D. Yang, Y. Wang, R. Pan, R. Chen, and Z. Chen. "State-of-health estimation for the lithium-ion battery based on support vector regression." Applied Energy 227 (2018): 273-283.
[20] J. Wei, G. Dong, and Z. Chen. "Remaining useful life prediction and state of health diagnosis for lithium-ion batteries using particle filter and support vector regression." IEEE Transactions on Industrial Electronics 65, no. 7 (2018): 5634-5643.
[21] Y. Zhang, R. Xiong, H. He, and M. G. Pecht. "Long short-term memory recurrent neural network for remaining useful life prediction of lithium-ion batteries." IEEE Transactions on Vehicular Technology 67, no. 7 (2018): 5695-5705.
[22] J. Wu, C. Zhang, and Z. Chen. "An online method for lithium-ion battery remaining useful life estimation using importance sampling and neural networks." Applied energy 173 (2016): 134-140.
[23] M. Kläs, A. M. Vollmer. "Uncertainty in Machine Learning Applications: A Practice-Driven Classification of Uncertainty." In International Conference on Computer Safety, Reliability, and Security, pp. 431-438. Springer, Cham, 2018.
[24] Y. Li, K. Liu, A. M. Foley, A. Zülke, M. Berecibar, E. N. Maury, J. V. Mierlo, and H. E. Hoster. "Data-driven health estimation and lifetime prediction of lithium-ion batteries: A review." Renewable and Sustainable Energy Reviews 113 (2019): 109254.
[25] K. Liu, Y. Li, X. Hu, M. Lucu, W.D. Widanalage, Gaussian Process Regression with Automatic Relevance Determination Kernel for Calendar Aging Prediction of Lithium-ion Batteries, IEEE Transactions on Industrial Informatics (2019). DOI: 10.1109/TII.2019.2941747.
[26] F. Li, J. Xu. "A new prognostics method for state of health estimation of lithium-ion batteries based on a mixture of Gaussian process models and particle filter." Microelectronics Reliability 55, no. 7 (2015): 1035-1045.
[27] Y. Peng, Y. Hou, Y. Song, J. Pang, and D. Liu. "Lithium-ion battery prognostics with hybrid Gaussian process function regression." Energies 11, no. 6 (2018): 1420.
[28] R. R. Richardson, C. R. Birkl, M. A. Osborne, and D. A. Howey. "Gaussian Process Regression for In Situ Capacity Estimation of Lithium-Ion Batteries." IEEE Transactions on Industrial Informatics 15, no. 1 (2019): 127-138.
[29] R. R. Richardson, M. A. Osborne, and D. A. Howey. "Gaussian process regression for forecasting battery state of health." Journal of Power Sources 357 (2017): 209-219.
[30] D. Yang, X. Zhang, R. Pan, Y. Wang, and Z. Chen. "A novel Gaussian process regression model for state-of-health estimation of lithium-ion battery using charging curve." Journal of Power Sources 384 (2018): 387-395.
[31] Y. Li, C. Zou, M. Berecibar, E. Nanini-Maury, J. C. W. Chan, P. V. D. Bossche, J. V. M., and N. Omar. "Random forest regression for online capacity estimation of lithium-ion batteries." Applied energy 232 (2018): 197-210.
[32] M. Fleischhammer, T. Waldmann, G. Bisle, B. I. Hogg, and M. W. Mehrens. "Interaction of cyclic ageing at high-rate and low temperatures and safety in lithium-ion batteries." Journal of Power Sources 274 (2015): 432-439.
[33] M. R. Palacín, A. de Guibert. "Why do batteries fail?." Science 351, no. 6273 (2016): 1253292.
[34] C. E. Rasmussen, and H. Nickisch. "Gaussian processes for machine learning (GPML) toolbox." Journal of machine learning research 11, no. Nov (2010): 3011-3015.
[35] D. Liu, J. Pang, J. Zhou, Y. Peng, and M. Pecht. "Prognostics for state of health estimation of lithium-ion batteries based on combination Gaussian process functional regression." Microelectronics Reliability 53, no. 6 (2013): 832-839.
[36] J. Zhao, L. Chen, W. Pedrycz, and W. Wang. "Variational Inference-Based Automatic Relevance Determination Kernel for Embedded Feature Selection of Noisy Industrial Data." IEEE Transactions on Industrial Electronics 66, no. 1 (2019): 416-428.
[37] J. Schmalstieg, S. Käbitz, M. Ecker, D. U. Sauer. "From accelerated aging tests to a lifetime prediction model: Analyzing lithium-ion batteries." In 2013 World Electric Vehicle Symposium and Exhibition (EVS27), pp. 1-12. IEEE, 2013.
[38] C. E. Rasmussen, "Gaussian processes in machine learning." In Summer School on Machine Learning, pp. 63-71. Springer, Berlin, Heidelberg, 2003.
[39] J. Yan, K. Li, E. Bai, X. Zhao, Y. Xue, and A. M. Foley. "Analytical Iterative Multistep Interval Forecasts of Wind Generation Based on TLGP." IEEE Transactions on Sustainable Energy 10, no. 2 (2018): 625-636.